\documentclass[letterpaper]{article} 
\usepackage{aaai24}  
\usepackage{times}  
\usepackage{helvet}  
\usepackage{courier}  
\usepackage[hyphens]{url}  
\usepackage{graphicx} 
\usepackage{natbib}  
\usepackage{caption} 
\usepackage{algorithm}
\usepackage{algorithmic}

\usepackage{newfloat}
\usepackage{listings}
\DeclareCaptionStyle{ruled}{labelfont=normalfont,labelsep=colon,strut=off} 
\floatstyle{ruled}
\newfloat{listing}{tb}{lst}{}
\floatname{listing}{Listing}
\usepackage{algorithm}
\usepackage{algorithmic}
\usepackage{multirow}
\usepackage{amsmath,amssymb,amsfonts}
\usepackage{mathtools}
\usepackage{commath}

\title{Few Shot Part Segmentation Reveals Compositional Logic for \\ Industrial Anomaly Detection}
\author{
    Soopil Kim\textsuperscript{\rm 1,3},
    Sion An\textsuperscript{\rm 1},
    Philip Chikontwe\textsuperscript{\rm 1},
    Myeongkyun Kang\textsuperscript{\rm 1,3}, \\
    Ehsan Adeli\textsuperscript{\rm 3},
    Kilian M. Pohl\textsuperscript{\rm 3},
    Sang Hyun Park\textsuperscript{\rm 1,2}
}
\affiliations{
    \textsuperscript{\rm 1}Robotics and Mechatronics Engineering, DGIST, Daegu, Korea\\
    \textsuperscript{\rm 2}AI Graduate School, DGIST, Daegu, Korea\\
    \textsuperscript{\rm 3}Stanford University, Stanford, CA 94305, USA\\
    \{soopilkim, shpark13135\}@dgist.ac.kr\\
    
}

\usepackage{bibentry}

\begin{document}

\maketitle

\begin{abstract}
Logical anomalies (\textit{LA}) refer to data violating underlying logical constraints \textit{e.g.,} the quantity, arrangement, or composition of components within an image. Detecting accurately such anomalies requires models to reason about various component types through segmentation. However, curation of pixel-level annotations for semantic segmentation is both time-consuming and expensive. Although there are some prior few-shot or unsupervised co-part segmentation algorithms, they often fail on images with industrial object. These images have components with similar textures and shapes, and a precise differentiation proves challenging. In this study, we introduce a novel component segmentation model for \textit{LA} detection that leverages a few labeled samples and unlabeled images sharing logical constraints. To ensure consistent segmentation across unlabeled images, we employ a histogram matching loss in conjunction with an entropy loss. As segmentation predictions play a crucial role, we propose to enhance both local and global sample validity detection by capturing key aspects from visual semantics via three memory banks: \textit{class histograms}, \textit{component composition embeddings} and \textit{patch-level representations}. For effective LA detection, we propose an adaptive scaling strategy to standardize anomaly scores from different memory banks in inference. Extensive experiments on the public benchmark MVTec LOCO AD reveal our method achieves 98.1\% AUROC in LA detection \textit{vs.} 89.6\% from competing methods.
\end{abstract}

\section{Introduction}
\noindent In industrial images, defects can be categorized into two main types: \textit{structural} and \textit{logical} anomalies \cite{bergmann2022beyond}. Structural anomalies (\textit{e.g.,} cracks and contamination) occur in localized regions often absent in normal data, whereas logical anomalies refer to data that does not adhere to underlying logical constraints, \textit{e.g.,} component composition and arrangement. Herein, effective detection requires the consideration of long-range dependencies within and across images. 

\begin{figure} [t]
\begin{center}
\includegraphics[width=1.0\linewidth] {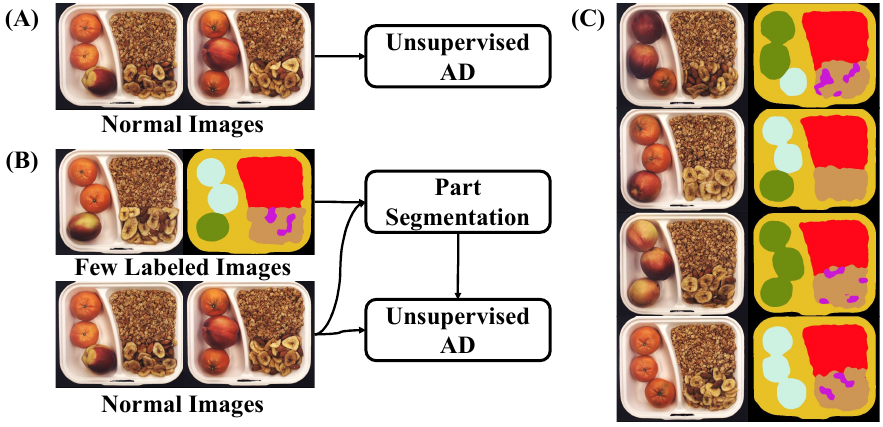}
\end{center}
\caption{Comparison of approaches at a conceptual level.
(A) The anomaly detection (AD) model is directly trained using images.
(B) Our proposed method guides part segmentation models using a few labeled samples to accurately segment components and then uses the segments for AD. (C) Examples of logical anomalies show the importance of semantically segmenting components for detection.}
\label{fig::teaser}
\end{figure}

Existing research on anomaly detection (AD) for industrial images has primarily focused on unsupervised approaches that aim to learn the distribution of normal data and detect outliers as anomalies. This has resulted in state-of-the-art models that have reported impressive scores exceeding 99\% \cite{roth2022towards}. This high score can be attributed to the nature of public benchmarks (\textit{e.g.,} MVTec AD \cite{bergmann2019mvtec} and MTD \cite{huang2020surface}), which predominantly comprises structural anomalies, resulting in models with much lower performance when targeting logical anomalies \cite{bergmann2022beyond}.

To address logical AD, current methods implicitly consider global dependencies among multiple components for effective detection, as described in Fig. \ref{fig::teaser}A. For example, \cite{bergmann2022beyond} proposed a hybrid feature reconstruction model, while \cite{tzachor2023set} introduced a histogram-based density estimation model. Despite these advancements, performance is constrained by the inability to accurately differentiate various components. For more accurate logical AD, it is essential to semantically segment the product's components, as they often exhibit similar features (\textit{e.g.,} peaches \textit{vs.} mandarins in Fig. \ref{fig::teaser}C). This task is closely related to co-part segmentation \cite{hung2019scops}, as normal samples' similar components follow pre-defined logics. However, existing unsupervised methods \cite{hung2019scops, gao2021unsupervised} often fail to precisely segment such components since they cannot distinguish similar features without relying on supervised guidance. 

A more effective approach could involve guiding part segmentation using a set of labeled images by employing manufacturers' prior knowledge about the individual elements required for product assembly as shown in Fig. \ref{fig::teaser}B. However, creating pixel-level annotations for numerous training images is a costly and labor-intensive task. While few-shot segmentation methods have made impressive advances to reduce the number of labeled samples \cite{wang2023seggpt, hong2022cost}, they equally fail to segment different parts that have similar textures or shapes.
To this end, we introduce a novel part segmentation model tailored to distinguish components in industrial images using few labeled images and several unlabeled images. Specifically, we utilize positional features for prediction and minimize a histogram matching loss for unlabeled images, ensuring each image maintains a consistent number of pixels per class. The combination of different losses enables the model to accurately segment elements across the images.

We integrate accurate part segmentation in our novel AD method, \textbf{PSAD} (\textbf{P}art \textbf{S}egmentation-based \textbf{A}nomaly \textbf{D}etection). Specifically, PSAD detects local and global dependencies of elements by relying on memory banks for \textit{class histograms}, \textit{component composition embeddings}, and \textit{patch-level representations}. To obtain a unified anomaly score from the different scaled outputs of the memory banks, we propose an adaptive strategy to re-scale anomaly scores using scores from training data. We evaluate the proposed method on a public dataset consisting of both logical and structural AD with five categories. We report results with higher AUROC compared to state-of-the-art not only in logical AD, but also in structural AD. Our contributions can be summarized as follows:
\begin{itemize}
\item We propose a novel anomaly detection method PSAD that employs 3 different memory banks by utilizing visual features and semantic segmentation.
\item We propose a new part segmentation method that is supervised by a limited number of labeled images with regularization using logical constraints shared across unlabeled images.
\item We propose an adaptive scaling method to aggregate anomaly scores with different scales.
\item Our method achieves state-of-the-art performance in both logical and structural anomaly detection.

\end{itemize}



\section{Related Works}

\noindent{\textbf{Anomaly Detection in Industrial Images}:}
In literature, existing anomaly detection (AD) methods often train models to first learn the distribution of normal data and then detect outliers as anomalies. These methods can be broadly categorized into \textit{reconstruction}, \textit{self-supervision}, and \textit{density estimation-based} models.

\textit{Reconstruction}-based methods learn to reconstruct normal input samples and determine the anomaly score based on the difference between inputs and the reconstruction \cite{lee2022cfa, liu2023diversity, tien2023revisiting}.
\textit{Self-supervised} methods create synthetic abnormal samples and use them to train a classifier. For instance, CutPaste \cite{li2021cutpaste} and DRAEM \cite{zavrtanik2021draem} generate abnormal samples for learning abnormality.
\textit{Density estimation} methods first extract features from normal samples using pre-trained models and then compare them with test sample features to compute anomaly scores \cite{roth2022towards, jiang2022softpatch, hyun2023reconpatch}.
We note that existing methods focus on utilizing local features since most benchmarks mainly contain structural anomalies rather than logical anomalies. 

Following the release of the first dataset comprising logical anomalies \cite{bergmann2022beyond}, several unsupervised methods have been proposed. 
GCAD \cite{bergmann2022beyond} trains local and global models that reconstruct pre-trained image features based on local and global dependencies. 
SINBAD \cite{tzachor2023set} extracts a set of orderless elements and randomly projects element features to compute a histogram, with anomaly scores obtained via density estimation. ComAD \cite{liu2023component} applied K-Means clustering on pre-trained features to segment multiple components within an image. However, performance was limited because a precise discrimination of different components is challenging.

We observe that product manufacturers are aware of the logical constraints on various components and this prior knowledge can be leveraged for AD. In this paper, we introduce a novel method PSAD using density estimation and semantic segmentation to precisely differentiate components. However, PSAD doesn't demand many labeled images due to our proposed few-shot segmentation method.

When using multiple anomaly scores and aggregating them, previous works simply add the scores \cite{tsai2022multi} or manually set hyper-parameters to scale them \cite{liu2023component}. However, these approaches may degrade performance when the multiple scores follow different distributions or the hyper-parameters are incorrectly set (Table \ref{table:abl_mb}). Even if \cite{bergmann2022beyond} attempted to normalize two distinct anomaly scores without defining hyper-parameters, they utilized a validation dataset to determine the statistics of these scores, potentially sacrificing valuable training data. Instead, we propose an adaptive scaling of the scores that solely relies on the training data by treating each sample as a test sample.

\begin{figure*} [t]
\begin{center}
\includegraphics[width=0.9\linewidth] {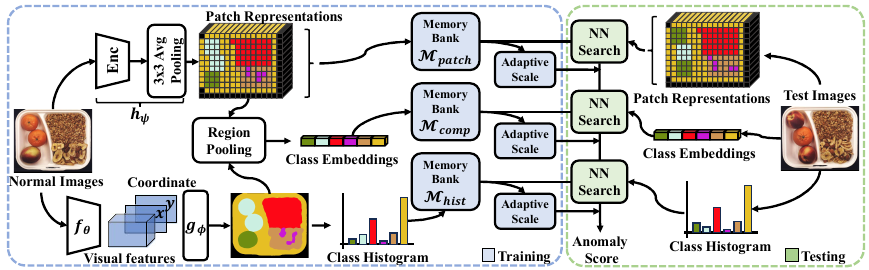}
\end{center}
\caption{Illustration of PSAD (Part Segmentation-based Anomaly Detection). During training depicted in the blue box, 3 different memory banks are constructed using normal images. The anomaly score of a test image is computed by finding its nearest neighbor (NN search) and adaptive scaling.}
\label{fig::ad}
\end{figure*}

\noindent{\textbf{Object Part Segmentation}:}
As part segmentation is vital for logical AD, one can train a supervised model for object part segmentation \cite{chen2014detect}. However, due to costly pixel labeling, unsupervised models \cite{sra2005generalized} that can learn arbitrary segmentation using a collection of unlabeled images are preferable.
\cite{hung2019scops} proposed an end-to-end segmentation model with pre-trained CNN features using semantic consistency and geometric concentration losses. Later, 
\cite{siarohin2021motion} and \cite{gao2021unsupervised} trained a segmentation model using a part-assembly procedure that reconstructs a target image by transforming parts of a source image. Though viable alternatives for industrial image segmentation, learning objectives based on geometric concentration or affine transformations are not generally applicable in industrial images since multiple objects can appear in distant positions (\textit{e.g.,} mandarins in `breakfast box' and hexagonal nuts in `screw bag' in MVTec LOCO dataset \cite{bergmann2022beyond}). In addition, models often under- or over-segment object parts as the labels are arbitrarily optimized. In this paper, we instead propose a new part segmentation model that can segment components in various industrial images using only few labeled samples.


\noindent{\textbf{Few Shot Semantic Segmentation}:}
Few-shot segmentation (FSS) has been proposed to overcome the data-hungry nature of deep learning models with different approaches employing generated or augmented images \cite{mondal2018few}, generative models \cite{tritrong2021repurposing, han2022leveraging, baranchuk2021label}, meta-learning \cite{hong2022cost, kim2023uncertainty}, transductive inference \cite{boudiaf2021few}, and foundation models \cite{wang2023seggpt}. 
In general, FSS models employing pre-trained generative models report good part segmentation, especially on several well-aligned images such as face or car. However, generative model training is challenging and requires several samples to guarantee good performance. Note that our method is closely related to the transductive approach RePRI \cite{boudiaf2021few} that uses a fixed pre-trained backbone and trains a pixel classifier with several regularization losses. During inference, only the classifier (prototype-based) is updated with few samples. While impressive, training is regularized by the initial segmentation which may be often noisy. Thus, we instead update the backbone and the classifier with a histogram matching loss to better utilize logical constraints shared across normal images.

\section{Methods}
\subsubsection{Problem Setting}
Unsupervised anomaly detection (AD) aims to train a model that can identify abnormal data from a set of normal data $\{X^1,..., X^{N_{train}}\}$ where $N_{train}$ is the number of data and their labels are all assigned as 0 (normal). The model is trained to distinguish between normal and abnormal test data, predicting labels as 0 (normal) or 1 (anomalous).

To detect logical anomalies (LA), accurate part segmentation has to be preceded. In this process, the class of each component is defined by the manufacturer as each normal image contains a predetermined number of specific parts appearing in predefined locations. Consequently, variations in the object's location may lead to different classes, even for the same object (e.g. `pushpins' and `splicing connectors' in Fig. \ref{fig::qual_fss}). Instance segmentation differs from semantic segmentation, as predefined class labels are not assigned to the instances. 
Since constructing pixel-level annotations of lots of images is labor-intensive, we assume that only a scarce number of labeled images  $\{X^{l,i}\in\mathbb{R}^{W\times H\times 3}, Y^{l,i} \in\mathbb{R}^{W\times H\times N_C}\}^{N_L}_{i=1}$ and a substantial set of unlabeled images $\{X^{u,j}\}^{N_U}_{j=1}$ are provided for training. Here, $N_L$, $N_U$, and $N_C$ represent the numbers of labeled images, unlabeled images, and classes, respectively. The model is optimized to reduce a combination of supervised losses for $X^l$ and unsupervised losses for $X^u$.

\subsubsection{Overview}
\begin{figure} [t]
\begin{center}
\includegraphics[width=1.0\linewidth] {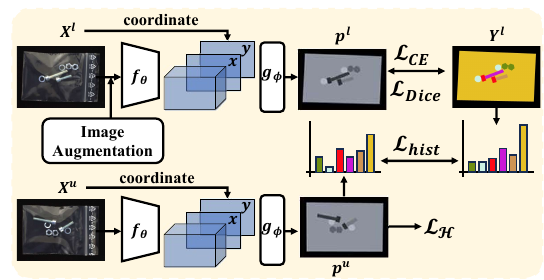}
\end{center}
\caption{Proposed part segmentation model that predicts segmentation utilizing visual and positional features. }
\label{fig::fss}
\end{figure}

Our proposed PSAD (Part Segmentation-based Anomaly Detection) consists of two parts: \textit{semantic part segmentation} and \textit{AD using part segmentation}. For part segmentation, we design a model to distinguish multiple components based on visual and positional features (Fig. \ref{fig::fss}). A visual feature extractor and a pixel classifier are jointly optimized with a few labeled images and logical constraints shared across numerous unlabeled images.
For AD with part segmentation, the segmentation model applied to normal samples is leveraged to construct three distinct memory banks (Fig. \ref{fig::ad}). In particular,
(1) a class histogram memory bank $\mathcal{M}_{hist}$ that records the quantity and arrangement of each component to assess the relative abundance of different components within the images.
(2) a component compositions memory bank $\mathcal{M}_{comp}$ that helps determine the validity of various compositions of components to identify anomalies that arise from unexpected or irregular component arrangements. Finally,
(3) a memory bank $\mathcal{M}_{patch}$ specifically designed for patch-level features to capture fine-grained details within the image.

From these memory banks, we generate three different anomaly scores, each of which has a different scale and distribution. To effectively compare and combine the scores, we perform adaptive scaling using statistics obtained from the training data to ensure scores can be reliably compared across different scales.

\subsection{Part Segmentation Using Limited Annotations}
The segmentation model consists of a feature extractor $f_{\theta}$ and a pixel classifier $g_{\phi}$, with $f_{\theta}(X)$ being a feature map having the same size as input $X$. Since the object's location is important, the pixels' coordinate $c\in\mathbb{R}^{W\times H\times 2}$ and $f_{\theta}(X)$ are concatenated as input for $g_{\phi}$. During training, given model prediction probability $p=g_{\phi}(f_{\theta}(X)\oplus c)$, parameters $\theta$ and $\phi$ are optimized via:
\begin{equation}
	\label{eq::total_loss}
	\mathcal{L} = \mathcal{L}_{Dice} + \lambda_1\mathcal{L}_{CE} + \lambda_2\mathcal{L}_{\mathcal{H}} + \lambda_3\mathcal{L}_{hist},
\end{equation}
where each $\lambda$ is a hyper-parameter. For labeled images $X^l$, our model relies on cross-entropy loss $\mathcal{L}_{CE}$ and dice similarity loss $\mathcal{L}_{Dice}$. Note that prediction on unlabeled images $X^u$ may be uncertain, especially with limited $X^l$. A common approach to handle this is by incorporating an entropy loss $\mathcal{L}_{\mathcal{H}}$ to reduce uncertainty \cite{wang2022semi}. Nonetheless, minimizing $\mathcal{L}_{\mathcal{H}}$ with only a few labeled images can lead to unexpected training outcomes and potentially degrade accuracy. To mitigate this, we propose a histogram matching loss $\mathcal{L}_{hist}$ to ensure consistency in segmenting each part with an equal number of pixels. We randomly select a label $Y^l$ from $\{Y^{l,i}\}^{N_L}_{i=1}$ and compare the class-level volume with predictions $p^u$ from unlabeled images:
\begin{equation}
	\label{eq::hist_loss}
	\mathcal{L}_{hist} = \frac{1}{N_C}\sum^{N_C}_{n=1} \norm{\frac{1}{WH}\sum_{w,h} Y^l_{w,h,n} - \frac{1}{WH}\sum_{w,h}p^u_{w,h,n}}
\end{equation}
While model parameters are updated to reduce uncertain predictions under the constraints from $\mathcal{L}_{CE}$, $\mathcal{L}_{Dice}$, and $\mathcal{L}_{hist}$, the model also learns consistent segmentation on numerous unlabeled images based on both the visual and positional similarity of each component.


\subsubsection{Handling Multiple Types of Products}
In industrial image datasets, products may be composed of various subtypes (e.g. `juice bottle' and `splicing connector' in the MVTec LOCO AD dataset). In such cases, it is necessary to ensure that $X^l$ and $X^u$ belong to the same product type for the comparison in $\mathcal{L}_{hist}$. To classify $X^u$ without human annotation, we compare unlabeled and labeled images in latent space and find the nearest labeled image for each unlabeled image. Specifically, we extract global-average-pooled features from the images using a pre-trained encoder before training the segmentation model. Each $X^u$'s type is determined as the type of the nearest labeled images in the latent space. Subsequently, $X^l$ and $X^u$ of the same type can be used together in $\mathcal{L}_{hist}$. As a result, our model can effectively handle datasets that contain multiple types of products.


\subsection{Anomaly Detection Using Part Segmentation}
Our proposed PSAD follows a density estimation approach \cite{defard2021padim} in which normal data features are stored in a memory bank $\mathcal{M}=\{e^k\}^{N_\mathcal{M}}_{k=1}$, with $N_\mathcal{M}$ denoting the number of elements in $\mathcal{M}$. To determine the anomaly score $s$ of a test sample $e_{test}$, we find its nearest neighbor among elements in $\mathcal{M}$: 
\begin{equation}
\label{eq::adscore}
 s = \mathop{\arg \min}\limits_{e \in \mathcal{M}} \norm{e_{test} - e}^2.
\end{equation}
Patch-level density-based methods \cite{roth2022towards, jiang2022softpatch} have proven to be effective in detecting structural anomalies by focusing on local features. However, logical anomalies often arise when multiple components appear together to form a single product or entity. 

\subsubsection{Class Histogram Memory Bank}
The first memory bank $\mathcal{M}_{hist}$ focuses on quantifying the number of components for each class through a class histogram. Given normal images and their corresponding segmentation, we construct a histogram that represents the distribution of pixels among different classes. The histograms are then stored in $\mathcal{M}_{hist}$ and respective anomaly scores are predicted using Eq.(\ref{eq::adscore}).

\subsubsection{Component Composition Memory Bank}
It is worth noting that solely relying on $\mathcal{M}_{hist}$ cannot verify whether the components are combined correctly or not. To address this, we introduce a component composition memory bank $\mathcal{M}_{comp}$ that stores feature compositions of different parts within an image. After a feature map $w = h_{\psi}(X)$ is extracted using a pre-trained encoder $h_{\psi}$, the segmentation map allows us to define a class embedding as an averaged feature vector of pixels belonging to each class. A concatenation of these class embeddings is saved in $\mathcal{M}_{comp}$ to effectively capture visual features of each component and their compositions within the image. The anomaly score is predicted using Eq.(\ref{eq::adscore}).

\subsubsection{Patch Representation Memory Bank}
Finally, we construct memory bank $\mathcal{M}_{patch}$ by storing patch-level representations to detect fine-grained features following established approaches \cite{defard2021padim, roth2022towards}: $\mathcal{M}_{patch} = \bigcup^{N_{train}}_{k=1} \{h_{\psi}(X^k)_{l}\}^{N_P}_{l=1}$, where $h_{\psi}(X^k)_{l}$ is the $l^{th}$ patch representations extracted from $X^k$ and $N_P$ denotes the number of patches. The anomaly score of $X_{test}$ is predicted as: $s = \mathop{\max}\limits_{e \in \{h_{\psi}(X_{test})_{l}\}^{N_P}_{l=1}} \mathop{\min}\limits_{e' \in \mathcal{M}_{patch}} \norm{e - e'}^2$.

\subsubsection{Aggregating Anomaly Scores of Different Scales}
Considering the distinct scales and distributions of the memory banks, it is essential to set appropriate hyperparameters for each anomaly score as arbitrarily configuring a single hyperparameter can negatively impact overall accuracy. To mitigate this, our solution is scaling $s$ based on the anomaly scores of training data in each memory bank. In particular, we derive a set of anomaly scores denoted as $S_{train} = \{s^1, ..., s^{N_{\mathcal{M}}}\}$ from the training data by treating each data point $e^k$ as a test sample. We then construct the memory bank using all other training samples excluding $e^k$ as follows: $s^k = \mathop{\min}\limits_{e \in \mathcal{M}, e \neq e^k} \norm{e^k - e}^2$.
In the context of $\mathcal{M}_{hist}$ and $\mathcal{M}_{comp}$, $e$ stands for a class histogram and a component composition embedding derived from a data sample, respectively. However, for $\mathcal{M}_{patch}$, $s^k$ is defined in a different way as multiple elements are saved from a data sample as: $s^k = \mathop{\max}\limits_{e \in \{h_{\psi}(X^k)_{l}\}^{N_P}_{l=1}} \mathop{\min}\limits_{e' \in \mathcal{M}_{patch}'} \norm{e - e'}^2$, where $\mathcal{M}_{patch}'=\bigcup^{N_{train}}_{m=1, m\neq k} \{h_{\psi}(X^m)_{l}\}^{N_P}_{l=1}$. 

We define a normalized anomaly score considering the statistics of $S_{train}$ as $\hat{s}_{\mathcal{M}} = s / \max \{s^1, ..., s^{N_{\mathcal{M}}}\}.$
This adaptive scaling approach improves accuracy and robustness in detecting anomalies. The final anomaly score is defined as a sum of three anomaly scores from different memory banks: 
$s=\hat{s}_{\mathcal{M}_{hist}}+\hat{s}_{\mathcal{M}_{comp}}+\hat{s}_{\mathcal{M}_{patch}}$, 
facilitating both structural and logical anomaly scoring. 


\begin{table*}[t!]
\footnotesize
\begin{center}
\begin{tabular}{clcccccccccc}
\hline
\multicolumn{1}{l}{} & \multicolumn{1}{l|}{Category} & PatchCore & RD4AD & DRAEM & ST & AST & GCAD & SINBAD & ComAD & SLSG & \bf{PSAD} \\ \hline
\multicolumn{1}{c|}{\multirow{6}{*}{LA}} & \multicolumn{1}{l|}{Breakfast Box} & 74.8 & 66.7 & 75.1 & 68.9 & 80.0 & 87.0 & 96.5 & 91.1 & - & \bf{100.0} \\
\multicolumn{1}{c|}{} & \multicolumn{1}{l|}{Juice Bottle} & 93.9 & 93.6 & 97.8 & 82.9 & 91.6 & \bf{100.0} & 96.6 & 95.0 & - & 99.1 \\
\multicolumn{1}{c|}{} & \multicolumn{1}{l|}{Pushpins} & 63.6 & 63.6 & 55.7 & 59.5 & 65.1 & 97.5 & 83.4 & 95.7 &  - &\bf{100.0} \\
\multicolumn{1}{c|}{} & \multicolumn{1}{l|}{Screw Bag} & 57.8 & 54.1 & 56.2 & 55.5 & 80.1 & 56.0 & 78.6 & 71.9 &  - &\bf{99.3} \\
\multicolumn{1}{c|}{} & \multicolumn{1}{l|}{Splicing   Connectors} & 79.2 & 75.3 & 75.2 & 65.4 & 81.8 & 89.7 & 89.3 & \bf{93.3} & - & 91.9 \\ \cline{2-12} 
\multicolumn{1}{c|}{} & \multicolumn{1}{l|}{\textbf{Average (LA)}} & 74.0 & 70.7 & 72.0 & 66.4 & 79.7 & 86.0 & 88.9 & 89.4 & 89.6 & \bf{98.1} \\ \hline

\multicolumn{1}{c|}{\multirow{6}{*}{SA}} & \multicolumn{1}{l|}{Breakfast Box} & 80.1 & 60.3 & 85.4 & 68.4 & 79.9 & 80.9 & \bf{87.5} & 81.6 & - & 84.9 \\
\multicolumn{1}{c|}{} & \multicolumn{1}{l|}{Juice Bottle} & 98.5 & 95.2 & 90.8 & \bf{99.3} & 95.5 & 98.9 & 93.1 & 98.2 & - & 98.2 \\
\multicolumn{1}{c|}{} & \multicolumn{1}{l|}{Pushpins} & 87.9 & 84.8 & 81.5 & 90.3 & 77.8 & 74.9 & 74.2 & \bf{91.1} & - & 89.8 \\
\multicolumn{1}{c|}{} & \multicolumn{1}{l|}{Screw Bag} & 92.0 & 89.2 & 85.0 & 87.0 & \bf{95.9} & 70.5 & 92.2 & 88.5 & - & 95.7 \\
\multicolumn{1}{c|}{} & \multicolumn{1}{l|}{Splicing   Connectors} & 88.0 & 95.9 & 95.5 & \bf{96.8} & 89.4 & 78.3 & 76.7 & 94.9 & - & 89.3 \\ \cline{2-12} 
\multicolumn{1}{c|}{} & \multicolumn{1}{l|}{\textbf{Average (SA)}} & 89.3 & 85.1 & 87.6 & 88.4 & 87.7 & 80.7 & 84.7 & 90.9 & 91.4 & \bf{91.6} \\ \hline
\multicolumn{1}{l}{} & \multicolumn{1}{l|}{\textbf{Average}} & 81.7 & 77.9 & 79.8 & 77.4 & 83.7 & 83.4 & 86.8 & 90.1 & 90.3 & \bf{94.9} \\ \hline
\end{tabular}
\end{center}
\caption{Performance comparison of the proposed model PSAD against state-of-the models on MVTec LOCO AD dataset. LA and SA denote logical and structural anomalies, respectively. Boldface represents the best score.}
\label{table:loco}
\end{table*}

\noindent{\textbf{Implementation Details}:}
We use a pre-trained Wide ResNet101 \cite{zagoruyko2016wide} for initializing the parameters of the segmentation model $f_{\theta}$. Among 4 convolutional blocks in $f_{\theta}$, features extracted from the first 3 blocks are resized to the size of input $X$ and concatenated to obtain $v\in\mathbb{R}^{W\times H\times(256+512+1024)}$. Labeled images were augmented following \cite{buslaev2020albumentations}. For training, we used an AdamW optimizer with a learning rate $0.001$ and batch size of 5 per iteration on an NVIDIA RTX A5000 GPU workstation. The model was first trained for 50 epochs using only $\mathcal{L}_{CE}$ and $\mathcal{L}_{Dice}$. After warming up with the supervised loss, the model is trained using Eq.(\ref{eq::total_loss}) for additional 50 epochs. As $\mathcal{L}_{Dice}$ is usually larger than the other losses, hyper-parameters $\lambda_1$, $\lambda_2$, and $\lambda_3$ were set as 10. 
$h_{\psi}$ has Wide ResNet101 as the visual feature encoder and a $3 \times 3$ average pooling operation following the setting of PatchCore \cite{roth2022towards} which is one of the state-of-the-art models proposed for detecting structural anomalies.

\begin{figure*} [t]
\begin{center}
\includegraphics[width=0.95\linewidth] {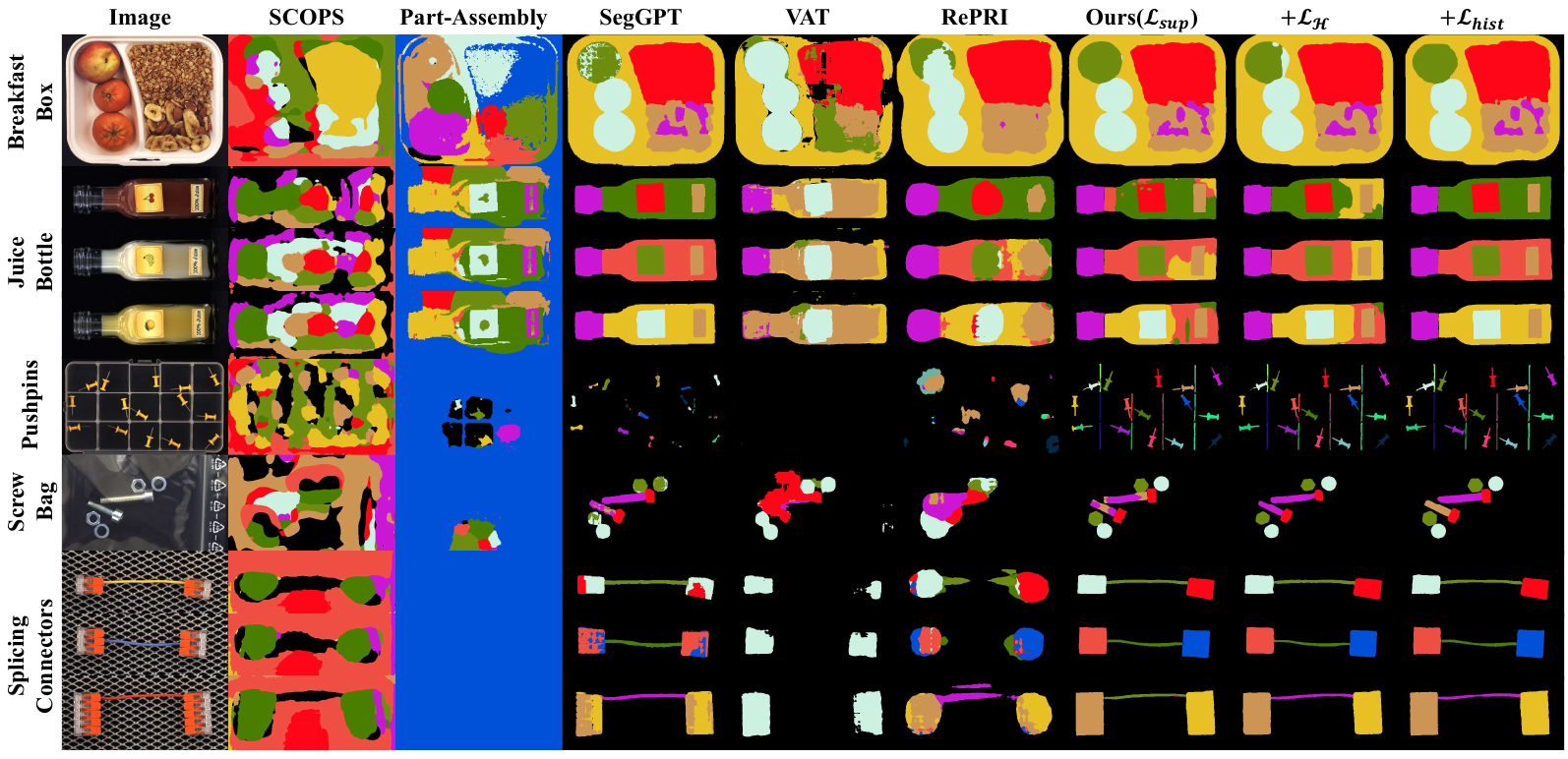}
\end{center}
\caption{Qualitative comparison of FSS models. $\mathcal{L}_{sup}$ ($=\mathcal{L}_{CE}+\mathcal{L}_{Dice}$) denotes a supervised loss for labeled images. For the unsupervised methods, such as SCOPS and Part-Assembly, we arbitrarily set the number of parts as 10.}
\label{fig::qual_fss}
\end{figure*}

\section{Experiments}

\noindent{\textbf{Experimental Setting}:}
We evaluated our method on MVTec LOCO AD dataset \cite{bergmann2022beyond}, the only benchmark for detecting logical anomalies to the best of our knowledge. This dataset consists of 5 categories (breakfast box/juice bottle/pushpins/screw bag/splicing connectors). For each category, 351/335/372/360/360 normal images were used for training and 275/330/310/341/312 images for testing following the setting of the comparison methods. Test data is categorized into good, structural anomaly (SA), and logical anomaly (LA). For the segmentation task, we used 5 labeled images since existing FSS models show more stable accuracy on the 5-shot setting. If the products have multiple types (\textit{e.g.,} 3 types within `juice bottle' and `splicing connectors'), we created a labeled image for each type, thereby employing a total of 3 labeled images.
State-of-the-art AD methods including PatchCore \cite{roth2022towards}, RD4AD \cite{deng2022anomaly}, DRAEM \cite{zavrtanik2021draem}, AST \cite{rudolph2023asymmetric}, ST \cite{bergmann2020uninformed}, ComAD \cite{liu2023component}, GCAD \cite{bergmann2022beyond}, SINBAD \cite{tzachor2023set}, and SLSG \cite{yang2023slsg} are used as comparison methods. The models are evaluated on SA and LA detection separately. We resized all images so that the number of pixels on the longer side among width and height is 512. The area under the ROC curve (AUROC) is used as a metric following previous works \cite{roth2022towards}.



\noindent{\textbf{Comparison with State-of-the-art Methods}:}
Table \ref{table:loco} lists the AUROC for LA and SA detection of methods trained and tested on MVTec LOCO AD dataset. Existing methods designed to focus on local features (e.g., PatchCore, RD4AD, DRAME, AST, ST) have lower scores in LA, as they can not capture the global dependencies among multiple components in the image, despite showing better scores in SA detection. While recent methods for detecting LA (e.g. ComAD, GCAD, SINBAD, SLSG) show improvements over those focusing on local features, they still fail to precisely distinguish between different components within the image. On the other hand, our proposed PSAD shows significant gains over others (i.e. +8.5 avg AUROC score across 5 categories in LA detection). 
Notably, we achieved a 99.3\% AUROC score in the 'screw bag' category, while the highest accuracy among the other methods in this category was only 80.1\%.
In addition, our proposed method also showed the best average score in SA detection. These results show that using semantic information can be beneficial for detecting both LA and SA. As a result, our proposed method obtained the best AUROC scores in both tasks.

\noindent{\textbf{Qualitative Comparison of FSS Methods}:}
In Fig. \ref{fig::qual_fss}, we show a qualitative comparison of different segmentation models on MVTec LOCO AD dataset. When we applied unsupervised co-part segmentation models SCOPS \cite{hung2019scops} and Part-Assembly \cite{gao2021unsupervised}, we obtained arbitrary segmentation results that can not discriminate between different components that are supposed to be segmented into different classes. In some cases, Part-Assembly failed to obtain proper results, as they focus on single-body objects. We also evaluated various state-of-the-art FSS models: a meta-learning-based model VAT \cite{hong2022cost}, a foundation model SegGPT \cite{wang2023seggpt}, and a transductive model RePRI \cite{boudiaf2021few}.
VAT and RePRI, each with ResNet-101 backbone pre-trained on PASCAL-$5^i$ dataset \cite{shaban2017one} showed poor performance in most cases. This is due to (1) frozen encoder during training, (2) encoder pre-trained on different domain data, and (3) models relying on inaccurate/noisy initial predictions. Though SegGPT showed relatively good results in some categories such as `juice bottle', it showed limited performance when multiple components have similar textures but different classes. For example in Fig. \ref{fig::qual_fss}, SegGPT fails to distinguish the left and right parts of `splicing connectors' and the short and long bolts in `screw bag', as they share similar textures. The limitations of existing FSS methods are mainly attributed to training models with existing segmentation datasets that do not necessitate considering position or length comparisons. As most meta-learning FSS models are focused on discriminating classes based on texture and shape, they equally show limited accuracy on industrial images which may have multiple similar components belonging to different classes. 

When we trained our model using only supervised losses for labeled images, we obtained better results in `pushpins' and `splicing connectors' but poor predictions in other categories despite using $\mathcal{L}_{\mathcal{H}}$. For example, it failed to discriminate between the short and long bolts in `screw bag'. However, when $\mathcal{L}_{hist}$ was employed, accurate segmentation results were obtained on various types of products. This shows that using $\mathcal{L}_{hist}$ with the other loss functions is more beneficial to obtain consistent segmentation by leveraging logical constraints.

\begin{table}[t]
\footnotesize
\centering
\begin{tabular}{l|cc}
\hline
\multicolumn{1}{c|}{Models} & \multicolumn{1}{c}{LA} & SA \\ \hline
SCOPS \cite{hung2019scops} & \multicolumn{1}{c}{82.5} & 90.2 \\
Part-Assembly \cite{gao2021unsupervised} & \multicolumn{1}{c}{80.3} & 85.6 \\ \hline
SegGPT \cite{wang2023seggpt} & \multicolumn{1}{c}{88.7} & 87.2 \\
VAT \cite{hong2022cost} & \multicolumn{1}{c}{79.2} & 87.8 \\
RePRI \cite{boudiaf2021few} & \multicolumn{1}{c}{83.6} & 88.4 \\ \hline
Ours ($\mathcal{L}_{sup}$) & \multicolumn{1}{c}{95.9} & 89.6 \\
Ours ($\mathcal{L}_{sup}+\mathcal{L}_{\mathcal{H}}$) & \multicolumn{1}{c}{96.3} & 90.0 \\
Ours ($\mathcal{L}_{sup}+\mathcal{L}_{\mathcal{H}}+\mathcal{L}_{hist}$) & \multicolumn{1}{c}{\bf{98.1}} & \bf{91.6} \\ \hline
\end{tabular}
\caption{Average AUROC scores of our proposed PSAD using different FSS models. }
\label{table:abl_fss}
\end{table}

\begin{table}[t]
\footnotesize
\centering
\begin{tabular}{cccc|cc}
\hline
$\mathcal{M}_{hist}$ & $\mathcal{M}_{comp}$ & $\mathcal{M}_{patch}$ & AS & LA & SA \\ \hline
\checkmark &  &  &  & 94.2 & 71.1 \\
 & \checkmark &  &  & 90.9 & 85.4 \\
 &  & \checkmark &  & 73.9 & 89.3 \\
\checkmark & \checkmark & \checkmark &  & 96.8 & 87.6 \\
\checkmark & \checkmark & \checkmark & \checkmark & \bf{98.1} & \bf{91.6} \\ \hline
\end{tabular}
\caption{Average AUROC scores of our proposed PSAD using different combinations of memory banks. `AS' stands for adaptive scaling.}
\label{table:abl_mb}
\end{table}

\begin{figure} [!t]
\begin{center}
\includegraphics[width=0.95\linewidth] {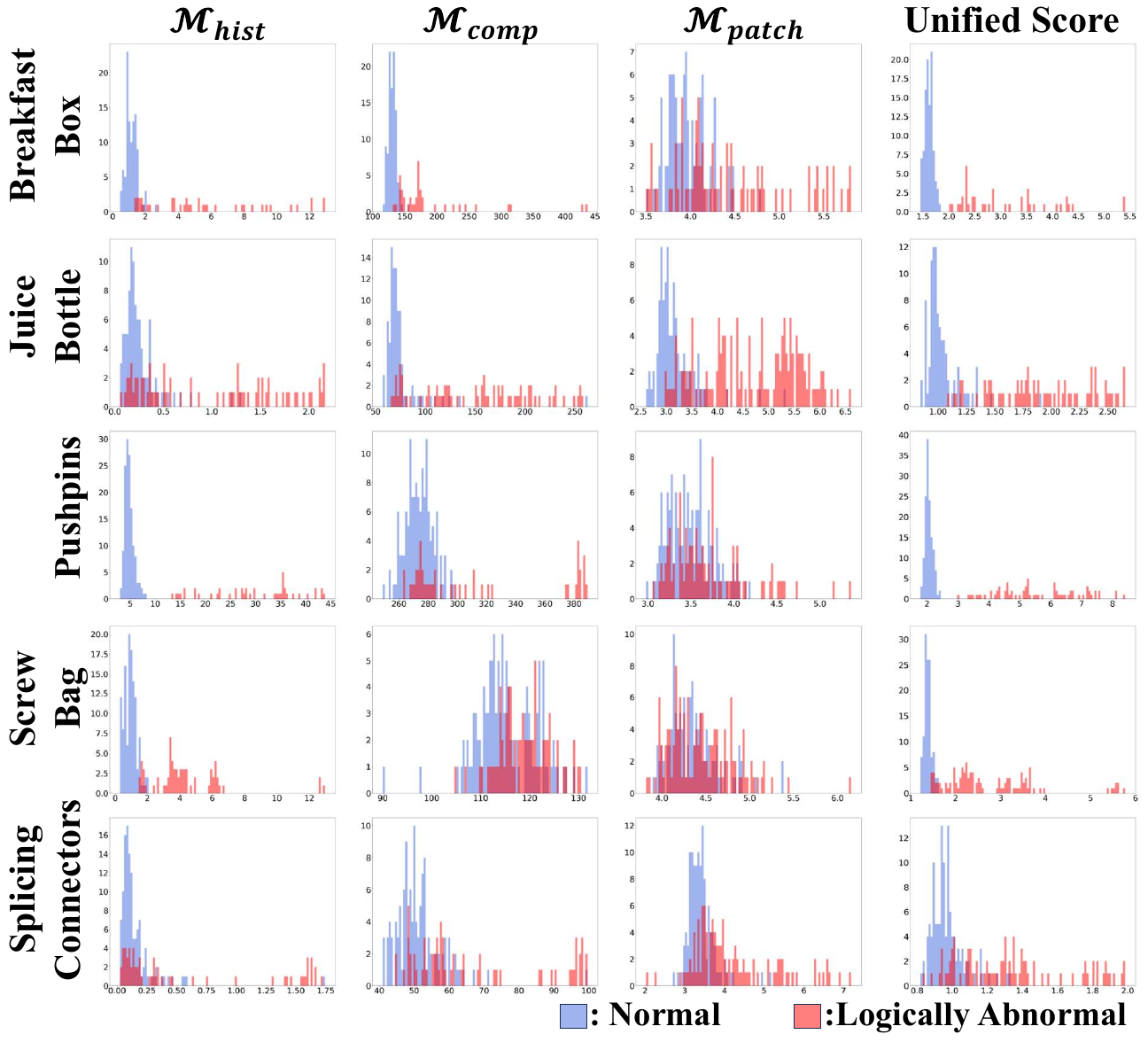}
\end{center}
\caption{Histogram visualizations of anomaly scores from different memory banks and the unified anomaly scores.}
\label{fig::qual_hist}
\end{figure}

\noindent{\textbf{Anomaly Detection Using Different Segmentation Models}:}
Based on the segmentation results, we also check whether accurate segmentation correlates with AD performance. Table \ref{table:abl_fss} shows the average AUROC scores of our methods using different segmentation models. FSS (SegGPT, VAT, and RePRI) and unsupervised models (SCOPS and Part-Assembly) showed low performance in LA detection, as they under-perform in segmenting some categories' data such as `pushpins'. Nevertheless, it is worth noting that their LA detection scores are higher than the score of our baseline PatchCore. This shows that leveraging even imperfect segmentation can be beneficial for LA detection. 

On the other hand, our model trained with $\mathcal{L}_{sup}$ ($=\mathcal{L}_{CE}+\mathcal{L}_{Dice}$) showed significantly improved scores even if it is not as accurate as our final model. This shows that our approach which jointly trains the encoder and classifier utilizing image augmentations and positional information is beneficial for segmenting industrial image data. When we use $\mathcal{L}_{\mathcal{H}}$ and $\mathcal{L}_{hist}$ together, we obtained further improved LA detection scores. Interestingly, the detection performance of SA is also enhanced with the utilization of more accurate segmentation results. These findings show the crucial role of accurate segmentation in achieving precise AD.

\noindent{\textbf{Effect of Various Memory Banks And Adaptive Scaling}:}
Table \ref{table:abl_mb} shows AD performance using different combinations of memory banks. When each memory bank is employed alone, $\mathcal{M}_{hist}$ and $\mathcal{M}_{comp}$ showed good performance on LA detection but poor performance on SA detection, whereas $\mathcal{M}_{patch}$ showed the opposite case. When we add these anomaly scores from 3 different memory banks without any scaling strategy, we obtained a better LA detection score and a degraded SA detection score. It is mainly attributed to varying scales of anomaly scores depending on memory banks. When we apply adaptive scaling, we obtained the best scores in both LA and SA detection.

Figure \ref{fig::qual_hist} illustrates histograms of anomaly scores obtained from various memory banks and the unified anomaly scores after adaptive scaling. It shows that the scale of anomaly scores varies across memory banks, implying the importance of scaling scores before aggregation.
Notably, the anomaly scores from the patch representation memory bank are poor due to its reliance on local features. Nevertheless, after normalizing each score and integrating them into a unified score, a clear discrimination between normal and abnormal samples was observed.
Overall, these findings highlight the significance of adaptive scaling in improving the effectiveness of AD using multiple memory banks.

\begin{table}[t]
\footnotesize
\centering
\begin{tabular}{c|cccc}
\hline
$N_{\mathcal{M}}$ & 100\% & 50\% & 25\% & 12.5\%  \\ \hline
Avg AUROC & 97.4 & 97.1 & 96.6 & 96.2 \\ \hline
\end{tabular}
\caption{AUROC in LA detection of our proposed PSAD using different numbers of normal images $N_{train}$. In this experiment, a combination of $\mathcal{M}_{hist}$ and $\mathcal{M}_{comp}$ is used. }
\label{table:less_data}
\end{table}

\noindent{\textbf{Logical Anomaly Detection Using Less Training Samples}:}
Table \ref{table:less_data} lists the AUROC of LA detection using varying numbers of unlabeled images. Despite a slight decrease in the average AUROC scores using less data, our approach still outperforms the other methods with the reduced dataset. This finding underscores the significance of accurate segmentation maps in enabling precise LA detection even with limited data. 


\section{Conclusion}
In this paper, we incorporate part segmentation into anomaly detection (AD) to detect logical and structural anomalies. To avoid constructing a large training dataset for segmentation, we propose a new segmentation model that utilizes a few labeled images and logical constraints shared across normal images. We also propose a novel AD method that involves constructing 3 distinct memory banks based on the segmentation. To generate a unified anomaly score from varying scales of anomaly scores, we introduce an adaptive scaling strategy. By doing so, our model could detect LA and SA, and yields substantial improvements with minimal effort required from users. As future few-shot segmentation models evolve to require fewer labeled images and produce better results, our AD model will achieve more enhanced performance with less effort from users. 


\section{Acknowledgments}
This work was supported by the DGIST R\&D program of the Ministry of Science and ICT of KOREA (22-KUJoint-02 and 21-DPIC-08) and the Digital Innovation Hub project supervised by the Daegu Digital Innovation Promotion Agency (DIP) grant funded by the Korea government (MSIT and Daegu Metropolitan City) in 2023 (DBSD1-01). 

\bibliography{aaai24}

\end{document}